\documentclass[journal,11pt]{elsarticle}
\usepackage{amsmath}
\usepackage{amssymb}
\usepackage{amsfonts}
\usepackage{graphicx}
\usepackage{fullpage}
\usepackage{graphics}
\usepackage{multirow}
\usepackage{epsfig}
\usepackage{setspace} 
\usepackage{epstopdf}
\usepackage{newfloat}
\usepackage{subcaption}
\usepackage{gensymb}
\usepackage[table]{xcolor}
\usepackage[normalem]{ulem}
\usepackage{algorithm}
\usepackage{algorithmic}
\usepackage{varioref}
\usepackage{hyperref}
\usepackage{bm}
\usepackage{array}
\usepackage{tcolorbox}

\definecolor{forestgreen}{rgb}{0.13, 0.55, 0.13}

\usepackage{blkarray}

\usepackage{lineno,hyperref}
\usepackage[normalem]{ulem}     
\newcommand\redout{\bgroup\markoverwith
{\textcolor{red}{\rule[0.5ex]{2pt}{0.8pt}}}\ULon}

\usepackage{soul}

\modulolinenumbers[5]

\newcolumntype{H}{>{\setbox0=\hbox\bgroup}c<{\egroup}@{}} 

\graphicspath{{./}{images/}}
\journal{Journal of \LaTeX\ Templates}

\begin{document}

\begin{frontmatter}

\title{Estimation of Correlation Matrices from Limited time series Data using Machine Learning}


\author[mymainaddress1]{Nikhil Easaw}
\author[mymainaddress2,mymainaddress3]{Woo Seok Lee}
\author[mymainaddress1]{Prashant Singh Lohiya}
\author[mymainaddress1]{Sarika Jalan}
\ead{sarika@iiti.ac.in}
\author[mymainaddress4]{Priodyuti Pradhan}
\corref{mycorrespondingauthor}
\ead{priodyutipradhan@gmail.com}
   
\address[mymainaddress1]{
Complex Systems Lab, Department of Physics, Indian Institute of
Technology Indore, Khandwa Road, Simrol, Indore-453552, India}

\address[mymainaddress2]{
Center for Theoretical Physics of Complex Systems, Institute for Basic Science (IBS), Daejeon 34126, Republic of Korea}

\address[mymainaddress3]{
1ST Biotherapeutics, Inc., Seongnam, 13493, Republic of Korea}

\address[mymainaddress4]{
Department of Computer Science and Engineering, Indian Institute of Information Technology Raichur, Karnataka - 584135, India}


\begin{abstract}
Correlation matrices contain a wide variety of spatio-temporal information about a dynamical system. Predicting correlation matrices from partial time series information of a few nodes characterizes the spatio-temporal dynamics of the entire underlying system. This information can help to predict the underlying network structure, e.g., inferring neuronal connections from spiking data, deducing causal dependencies between genes from expression data, and discovering long spatial range influences in climate variations. Traditional methods of predicting correlation matrices utilize time series data of all the nodes of the underlying networks. Here, we use a supervised machine learning technique to predict the correlation matrix of entire systems from finite time series information of a few randomly selected nodes. The accuracy of the prediction validates that only a limited time series of a subset of the entire system is enough to make good correlation matrix predictions. Furthermore, using an unsupervised learning algorithm, we furnish insights into the success of the predictions from our model. Finally, we employ the machine learning model developed here to real-world data sets.
\end{abstract}

\begin{keyword}
\texttt{Time series data\sep Correlation matrix\sep Non-linear dynamics\sep Machine learning \sep Complex networks 
}
\end{keyword}

\end{frontmatter} 


\section{Introduction}
Machine learning (ML) has been applied in diverse areas of physical sciences ranging from condensed matter to high-energy physics to complex systems. In complex systems, neural network-based ML techniques have been used in predicting amplitude death \cite{amplitude_death}, the anticipation of synchronization \cite{Anticipating_synchronization}, phase transitions in complex networks \cite{ml_phase_transition}, time series prediction \cite{ghosh2021reservoir}, etc. In particular, forecasting time series data has attracted interest from the scientific fraternity due to its diverse applications in real-world dynamical systems like price prediction in stock markets and the EEG time series analysis of brain \cite{PhysRevLett.80.5019,lainscsek2013non,gerster2020fitzhugh}. However, predicting a time series data point of a dynamical system has many limitations \cite{ModelFreePrediction, LongTermPrediction}. Since every data point in a time series is a function of the previous time steps, the error in predicting future data points compounds over time. To avoid prediction error, the correlation matrix of the time series is preferred over a direct prediction of the future time series data points. A correlation matrix of a given multivariate time series data set is advantageous in several practical, real-world scenarios \cite{schindler2007assessing, PropertiesOfCrossCorrelations}. Spatio-temporal correlation patterns characterize the dynamics of a system \cite{muller2005detection}. For instance, by considering fMRI or MEG signals from several brain regions as time series data, one can construct the corresponding correlation matrix, which can then be used to extract the adjacency matrix by setting a threshold value \cite{4,6,7}.  

In most cases, one calculates average correlation matrices of a given time series data. A correlation matrix of time series data may vary depending on the length of observations and temporal position. Two well-known methods of estimating a true correlation matrix are; (i) the maximum likelihood estimation (MLE) and (ii) the graphical least absolute shrinkage and selection operator method (GLASSO). The MLE method first assumes a sample correlation matrix from a Gaussian distribution that is iteratively corrected to estimate an actual correlation matrix by maximizing the likelihood of observing the given time series data. The GLASSO method \cite{8} is an extension of the MLE method for those cases where MLE can not be applied, for example, if the dimensionality of the Gaussian distribution is higher than the available number of observation samples. Furthermore, a prerequisite of these techniques is to have information on time series data of all the network nodes, whereas, in real-world systems, often time series information of limited nodes are available. Therefore, these methods stall modeling cases where the number of time series is much lesser than the number of nodes forming the corresponding system.

\begin{figure*}
    \centering
    \includegraphics[width=5.8in,height=4in]{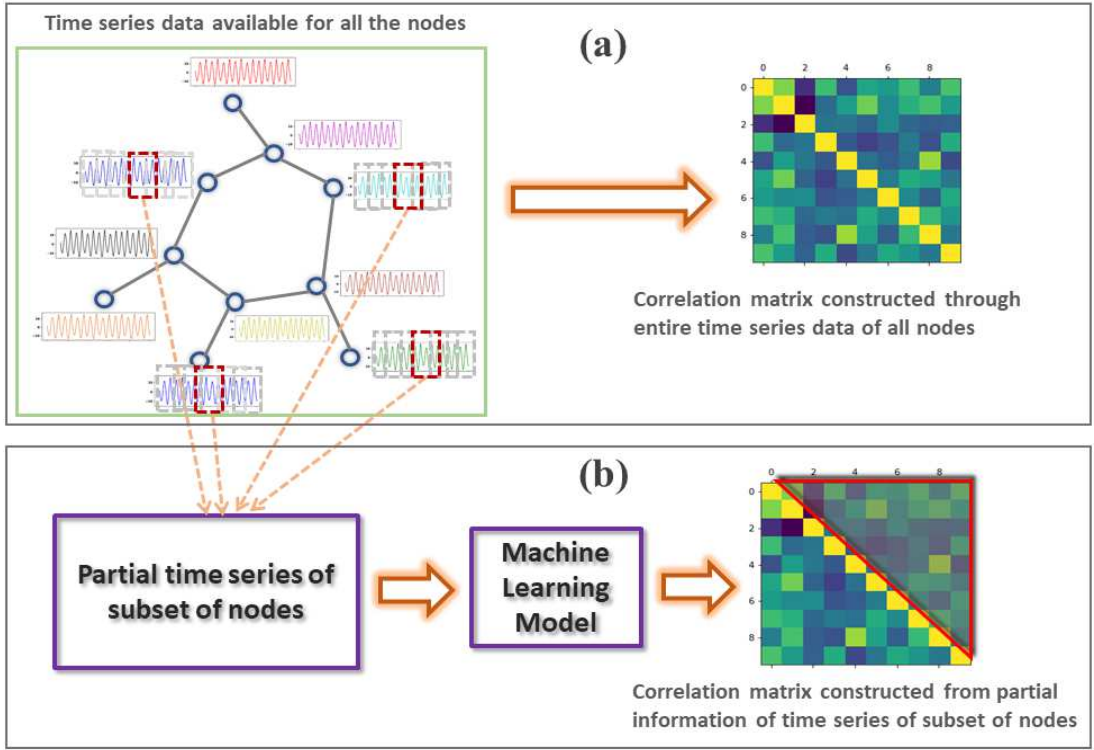}
    \caption{Schematic representation of (a) traditional and (b) ML model structure for constructing correlation matrix from time series data on networks. Input to the traditional model is the time series data of all the nodes, and output is the correlation matrix for all the nodes. For the ML model, input is partial time series from a few nodes in the shape of a time series window, and output is the predicted upper triangular part of the correlation matrix for all the nodes.}
    \label{schematic_diagram}
\end{figure*}

We develop an ML framework to reconstruct a full correlation matrix from partial time series data of a few nodes (Fig. \ref{schematic_diagram}). By considering different dynamical systems, we demonstrate that the supervised learning method can predict the correlation matrix from a few nodes' limited time series data. Furthermore, by analyzing the mean square error (MSE) between the true and the predicted correlation matrices, we confirm that only a limited time series data for a subset of nodes is enough to accomplish good predictions. The correlation matrices are predicted by considering time series data associated with a given network's higher-degree or lower-degree nodes. The prediction accuracy for both cases is the same, indicating that the degree of the available nodes associated with a time series data does not impact the correlation matrix prediction. Furthermore, we use an unsupervised learning algorithm (UMAP) to provide insights into our forecasts by visualizing the true and predicted correlation matrices as points in 2D space. Finally, we use real-world neurological data sets to validate our model.   

The article is organized as follows: Section 2 discusses the graphs, dynamical, and ML models. It also contains the notations and definitions followed in the paper. Section 3 illustrates the time series data generation method, results, and analysis. Finally, section 4 summarizes our study.

\section{Preliminary}
Consider an un-directed graph or network, $\mathcal{G} =\{V,E, {\bf X}_{V}(t)\}$ where $V=\{v_1,\ldots,v_N\}$ is the set of vertices (nodes), $E = \{(v_i, v_j) | v_i, v_j \in V \}$ is the set of edges (connections) which contains the unordered pairs of vertices and ${\bf X}_{V}(t)$ is the dynamic state representing the time series data on the nodes. We denote the adjacency matrix corresponding to $\mathcal{G}$ as ${\bf A} \in \mathbb{R}^{N \times N}$ which is defined as $A_{ij} = 1$ if nodes $i$ and $j$ are connected, and $0$ otherwise. The $|V|=N$ and $|E|=M$ represent the number of nodes and edges in $\mathcal{G}$, respectively. The number of edges linked to a particular node $v_i$ is referred to as its degree and denoted by $k_i = \sum_{j=1}^{N} a_{ij}$. We refer degree sequence of $\mathcal{G}$ as $k_1,k_2,\ldots,k_N$ and any $n$ subset of higher (HD) or lower (LD) degree nodes as $\{k_i\}_{i=1}^{n}$. We refer to the minimum degree node as $k_{min} = \min_{1\leq i \leq N} k_i$ and maximum degree node or the hub node as $k_{max} = \max_{1\leq i \leq N} k_i$. The average degree of the network is denoted by $\langle k \rangle$ = $\frac{1}{N}\sum _{i=1}^{N} k_{i}$.

We use two random graph models, the Erd\H{o}s-R\'enyi (ER) and the Scale-Free (SF) networks, to model the coupled dynamical systems \cite{barabasi1999emergence}. The ER random network or the graph-valued random variable with the parameters is denoted by $\mathcal{G}^{ER}(N,p)$ where $N$ is the number of nodes and $p$ is the edge probability \cite{blum2020foundations}. The existence of each edge is statistically independent of all other edges. When we refer ``the graph $\mathcal{G}^{ER}(N, p)$," we mean one {\em realization} of the random variable with mean degree $\langle k \rangle$ and generated as follows. Starting with $N$ number of nodes, connecting them with a probability $p=\langle k \rangle / N$. The ER random network realization thus generated will have a Binomial degree distribution. The SF networks ($\mathcal{G}^{SF}$) generated using the Barab{\' a}si-Albert model follows a power-law degree distribution \cite{barabasi1999emergence}. 

\subsection{Dynamical Models}
\begin{figure}[tbh]
    \centering
\includegraphics[width=6in,height=3.2in]{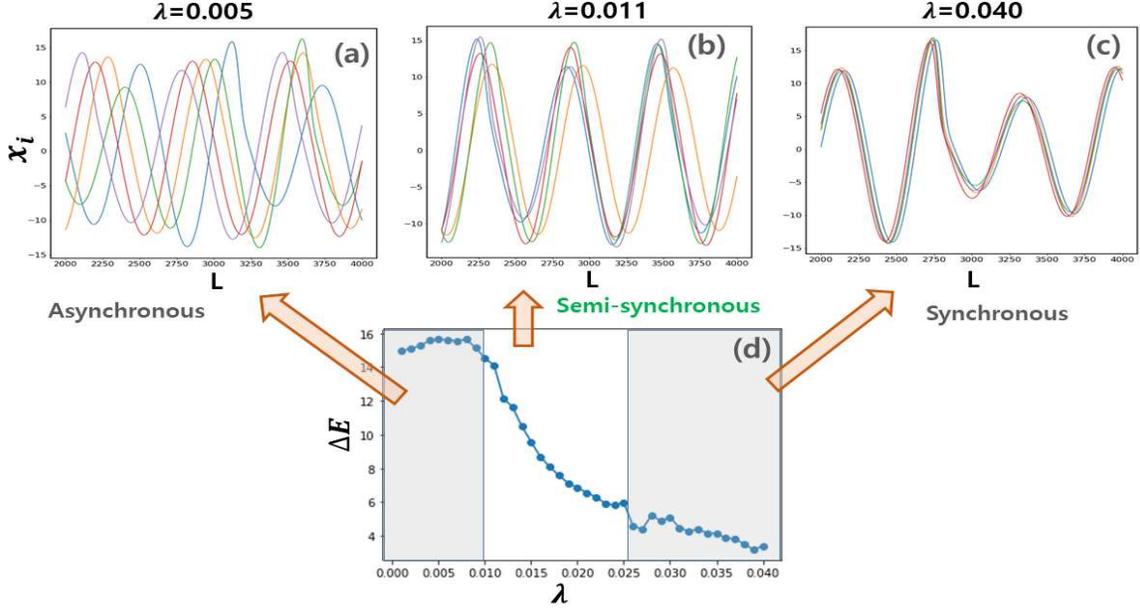}
   \caption{The behavior of time series data generated from the R{\" o}ssler oscillators on ER random networks when varying the coupling strength ($\lambda$). (a) At lower values of $\lambda$, the system is in an asynchronous state, i.e., generated time series data are uncorrelated. (b) After that, time series data are neither too correlated nor uncorrelated (semi-synchronous state), and (c) at higher values of ${\bf \lambda}$, time series data are highly correlated (synchronized state). The prediction becomes trivial if the coupling strength leads to a highly correlated or uncorrelated time series data set. Therefore, we choose $\lambda$ values such that generated time series data lie in the semi-synchronous state. (d) We quantify the level of correlation ($\Delta {\bf E}$) for the oscillators using global synchronization error (Eq. \ref{GSE}).}
    \label{GSE_error}
\end{figure}
We consider chaotic R{\" o}ssler and FitzHugh-Nagumo neuronal oscillators to model the dynamical evolution of nodes \cite{rosenblum1996phase, hodgkin1952quantitative}. The coupled dynamics of the nodes on a given graph generate time series data for the entire system. Dynamical evolution of the state 
of each node in the network is modeled by the R{\" o}ssler oscillator \cite{rosenblum1996phase} as follows
\begin{equation}\label{rossler_equations}
\begin{split}
  \dot{x_i} & = -\omega_iy_i-z_i+\lambda_{R}\sum_{j=1}^{N}A_{ij}(x_j-x_i)\\ 
  \dot{y_i} & = \omega_i x_i + a y_i \\ 
\dot{z_i} & = b + z_i(x_i - c)
\end{split}
\end{equation} 
where $x_i$, $y_i$, $z_i$ for $i=1,\ldots,N$ are the dynamical state variables and $\omega_i$ is the natural frequency of $i^{th}$ node which is drawn from a normal distribution with mean $1$, variance $0.03$. We choose other parameters for the chaotic region as $a = 0.15$, $b = 0.2$, and $c=10$ \cite{rosenblum1996phase}. Here, $A_{ij}$ represents a connection between nodes $i$ and $j$ of $\mathcal{G}$, and $\lambda_{R}$ denotes the overall coupling strength between the connected nodes. 

Next, we consider the FitzHugh-Nagumo (FHN) model, derived from the works of the Hodgkin-Huxley neuronal dynamical model \cite{hodgkin1952quantitative}. Many variations of the original FHN model have been developed since it was first introduced. In the present study, we consider the FHN model govern by the following equations \cite{su2014identifying}:
\begin{equation}\label{FHN_equations}
\begin{split}
 \dot{x_i} &= \frac{1}{\delta} [x_i(x_i-a)(1-x_i)-y_i]+\lambda_{F}{\sum_{j=1}^N}{A_{ij}}(x_j-x_i)\\
 \dot{y_i}&=x_i-y_i-b+S(t)
\end{split}
\end{equation}
where $x_i$ indicates the membrane potential and $y_i$ stands for the recovery variable of the $i^{th}$ node. We choose the frequency ($\omega_i$) of the driving signal  ($S(t) = r\sin \omega_i t$) from the normal distribution having mean $15$ and variance $0.001$. We select other parameters of the oscillators for the chaotic region as $a = 0.42$, $b = 0.15$, $\delta = 0.005$, and $r = 0.2$. Here, $\lambda_{F}$ denotes the overall coupling strength. The important parameter for our purpose is the coupling strengths ($\lambda_{R}$ and $\lambda_{F}$) and the network realizations encoded in ${\bf A}$. 

We vary $\lambda_{R}$, $\lambda_{F}$, and ${\bf A}$ to generate time series data sets having different correlation matrices. Fig. \ref{GSE_error}(a-c) shows three different behaviors of the generated time series data as asynchronous, semi-synchronous, and synchronous. We choose the semi-synchronous region for our study (Fig. \ref{GSE_error}(d)). Furthermore, we quantify the strength of correlation of the time series data of the nodes using the global synchronization error ($\Delta {\bf E}$) as \cite{boccaletti2018synchronization}
\begin{equation}\label{GSE}
 {\bf \Delta E}=\frac{1}{N}\sum_{i=1}^{N} \Delta E_i\text{, } \Delta E_i =\frac{1}{LN}\sum_{j=1}^{N}\sum_{k=t}^{t+L}||\bm{x}_{ki}-\bm{x}_{kj}||_2 \end{equation}
where $N$ is the number of nodes, $L$ is the length of time series data after the initial transient time (t). Here, $\Delta {\bf E}_i$ gives the error as the Euclidean distance from $i^{th}$ to $j^{th}$ node's oscillator for the $k^{th}$ time steps, $||\bm{x}_{ki}-\bm{x}_{kj}||_2=\sqrt{(x_{i}-x_{j})^2+(y_{i}-y_{j})^2+(z_{i}-z_{j})^2}$, where $x_i$, $y_i$, $z_i$ are the state variables of the R\"ossler oscillator. Similarly, we calculate $\Delta {\bf E}$ for the FHN oscillators (Fig. S1). 

\subsection{Machine Learning Algorithms}
\begin{figure*}[tbh]
\centering
\includegraphics[width=6.5in, height=2.8in]{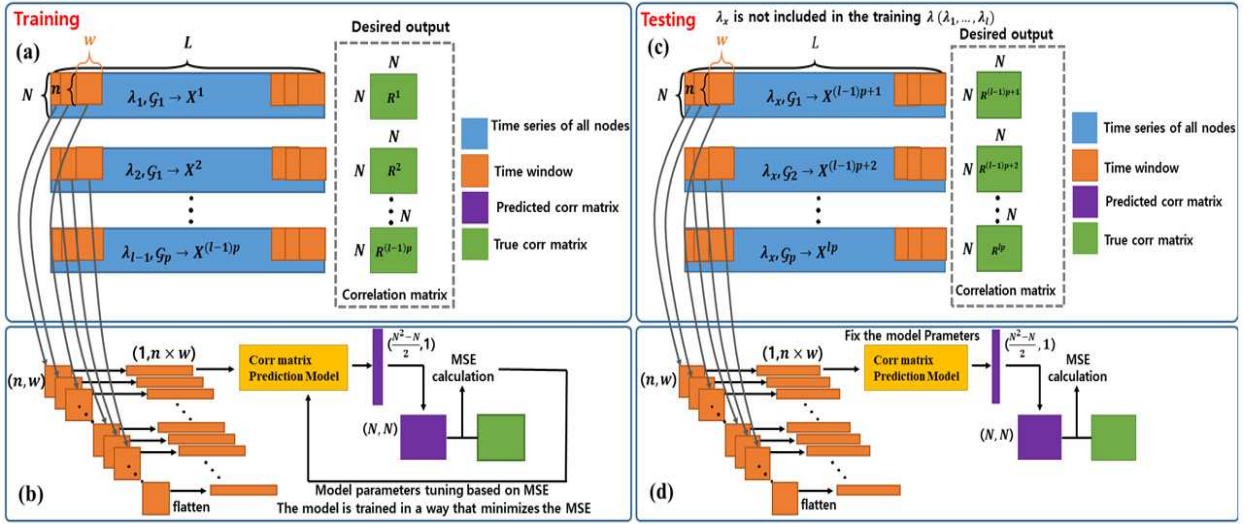}
\caption{Illustrate the training and testing phase of the ML model. For each $\lambda_i$, $\mathcal{G}_j$ ($1 \leq i\leq l$, $1 \leq j\leq p$), we have corresponding time series data set and true correlation matrix (${\bf X}^{k} \in \mathbb{R}^{N \times L}$, ${\bf R}^{k} \in \mathbb{R}^{N \times N}$, $1 \leq k\leq lp$). (a) To use the ML model, we generate time windows of size $n \times w$ ($n<N, w<L$) from the time series data set ${\bf X}^{k}$ and assign the corresponding true correlation matrix as the desired output. (b) The time windows and correlation matrices are flattened to train the ML model. For the correlation matrix, only the upper triangular part ($\frac{N^2-N}{2},1$) is considered. The correlation matrix predicted from a specific time window is used for updating the model parameters during the training phase by calculating the MSE between the model output (predicted correlation matrix) and desired output (true correlation matrix). (c, d) During the test phase, for a given window as input, the ML model can predict the upper triangular part of the correlation matrix as output. All parameters are updated during the training phase and are fixed during the test phase.}
\label{schematic_train_test}
\end{figure*}
We use a supervised learning algorithm to predict the correlation matrices. The ML model used to predict the correlation matrix is the feed-forward neural network referred to as multi-layer perceptron (MLP) \cite{goodfellow2016deep}. The architecture of this model contains one input layer, two hidden layers, and one output layer. 
A layer comprises several neurons, and neurons in the adjacent layers are connected. We adopt the SELU (scaled exponential linear unit) as a basic activation function except for the output layer \cite{klambauer2017self}. We use sigmoid as the output activation function when the desired output lies between $0$ and $1$, and $\tanh$ when it is between $-1$ and $1$. The number of neurons in the first and second hidden layers was set to $1225$ and $5041$, respectively, using hyper-parameter optimization techniques called HyperOpt \cite{bergstra2013hyperopt}. The relationship between the input ($a_{j}^{(\ell-1)}$) and the output ($a_{i}^{(\ell)}$) of a layer can be given by,
\begin{equation} \label{eq.FCN_layer}
z^{(\ell)}_{i}=\sum^{K}_{j=1}w^{(\ell-1)}_{ji}a^{(\ell-1)}_{j},\;\;\; 
a^{(\ell)}_{i} = \textrm{SELU}(z^{(\ell)}_{i})
\end{equation}
where $w_{ij}$ is a weighted connection between the $j^{th}$ neuron of the $(\ell-1)^{th}$ layer and $i^{th}$ neuron of the $(\ell)^{th}$ layer, with $K$ denoting the number of neurons in $(\ell-1)^{th}$ layer. Value of $a_{j}^{(\ell)}$ indicates output of $j^{th}$ neuron in the $\ell^{th}$ layer. The neural network receives an input ($a^{(0)}$) and generates an output ($a^{(L)}$) through the above propagation rule. In our case, $a^{(0)}$ is the input time series window, and $a^{(L)}$ is the upper triangular part of the predicted correlation matrix (Fig. ~\ref{schematic_diagram}). Training a neural network means finding $w_{ij}$ which can give us the desired output (upper triangular part of ${\bf R}\in \mathbb{R}^{N \times N}$ denoted as $\bm{r}\in \mathbb{R}^{m}$) for a given input window by reducing the difference between the neural network output $a^{(L)}$ and the desired output ($\bm{r}\in \mathbb{R}^{m}$), where $m=\frac{N^2-N}{2}$. We define this difference  and call it a loss function as 
\begin{equation}\label{eq.cost_fnc}
    \mathcal{L} = \frac{1}{m}\sum_{i=1}^{m}{(r_i-a^{(L)}_i)^2}
\end{equation}
We use Adam (Adaptive Moment Estimation) algorithm to minimize the loss function \cite{kingma2014adam}.
Furthermore, we use an unsupervised ML approach (UMAP) to gain more insights into the predicted correlation matrices by projecting and visualizing them in lower-dimensional subspace. The UMAP (Uniform Manifold Approximation and Projection) is a manifold learning technique for dimension reduction \cite{fujiwara2020visual}. The method preserves the local and global structures of the data set. Data having similar structures or features are clustered together in low dimensions. The use of UMAP here is two-fold -- (a) it provides a visual understanding of the predicted correlation matrices with the true correlation matrices, and (b) it helps us to select the appropriate training data set to get a meaningful prediction (SI sec. 4). 

\section{Methods and Results}

We predict correlation matrices for cases that closely match real-world scenarios. A dynamic system with a fixed number of nodes can undergo two types of changes (1) coupling strength between nodes can either increase or decrease. (2) a small structural change with rearrangement of links between nodes. We prepare time series data sets to match these cases. 

\subsection{Time series data generation and representation} 

To prepare data sets, we use two different dynamical models (R{\" o}ssler and FHN oscillators) with $l$ different coupling strengths ($\lambda_1,\lambda_2,\ldots, \lambda_l$) on two different model networks (ER and SF) each of having $p$ different realizations ($\mathcal{G}_1,\mathcal{G}_2,\ldots, \mathcal{G}_p$). Hence, we have $d=lp$ different times-series data sets ($\{{\bf X}^1,{\bf X}^2,\ldots,{\bf X}^{d}\}$) and its associated true correlation matrices ($\{{\bf R}^1,{\bf R}^2,\ldots,{\bf R}^{d}\}$) for each of the network and dynamical models, respectively (Fig. \ref{schematic_train_test}). To generate time series data sets, we numerically solve the coupled dynamical systems by varying coupling strength and the realization of model networks. Although on each node of the network, there are three dynamical state variables ($x_i$, $y_i$, and $z_i$) for the R{\" o}ssler (Eq. \ref{rossler_equations}) and two dynamical state variables ($x_i$ and $y_i$) for the FHN oscillators (Eq. \ref{FHN_equations}), we consider only the time evolution of $x_i$ variables of the oscillators as the time series data sets. After removing the transient part of the time evolution of state variables, we consider time series data sets and construct the corresponding true correlation matrices. For a $N$ size network, we will have $N$ variable time series data, each having length $L$ and stored as a time series matrix ${\bf X}^{k}\in \mathbb{R}^{N\times L}$, $1 \leq k \leq d$ as 
\begin{equation}\nonumber
\begin{gathered}
{\bf X}^{k}
    =\begin{pmatrix}
x_{11}^{k}  & x_{12}^{k}  \dots & x_{1L}^{k}\\
x_{21}^{k}  & x_{22}^{k}  \dots & x_{2L}^{k}\\
\vdots  & \vdots   \ddots & \vdots\\
x_{N1}^{k}  & x_{N2}^{k}   \dots & x_{NL}^{k}
\end{pmatrix},\;
{\bf R}^{k}=
\begin{pmatrix}
r_{11}^{k}  & r_{12}^{k} \dots & r_{1N}^{k}\\
r_{21}^{k}  & r_{22}^{k} \dots & r_{2N}^{k}\\
\vdots  & \vdots \ddots & \vdots\\
r_{N1}^{k}  & r_{N2}^{k} \ddots & r_{NN}^{k}
\end{pmatrix},\;
\bm{r}^{k}=
\begin{pmatrix}
r_1 \leftarrow r_{12}^{k} \\
\vdots\\
r_i \leftarrow r_{1N}^{k}\\
r_{i+1} \leftarrow r_{23}^{k}\\
\vdots \\
r_m \leftarrow r_{N-1N}^{k}
\end{pmatrix}
\end{gathered}
\end{equation}
where each row of the matrix represents individual nodes, and the time evolution of each node is recorded in columns. Here, $x_{ij}^{k}$ represents the time series information of the $i^{th}$ node at the $j^{th}$ time step for $k^{th}$ time series data set. We measure the influence of one node on another as the correlation between their time series data. We evaluate the correlation between a pair of time series data of nodes using Pearson correlation coefficient (Spearman) and stored in a  matrix (${\bf R}^{k} \in \mathbb{R}^{N \times N}$) referred to as {\em true} correlation matrix, where $r_{ij}^{k}$ represents the correlation between the time series of $i^{th}$ and $j^{th}$ nodes  \cite{pearson1}. Considering the symmetry of the correlation matrix, we take only the upper triangular part (diagonal components were also excluded) and construct a column vector, $\bm{r}^{k}\in \mathbb{R}^{m \times 1}$, where $m=\frac{N^2-N}{2}$ for the ML model output 
(Figs. \ref{schematic_diagram}(b) and \ref{schematic_train_test}(b)). Therefore, we assign $m$ neurons for the output layer. 
\begin{figure}[thb]
\centering
\includegraphics[width=6.5in, height=3.2in]{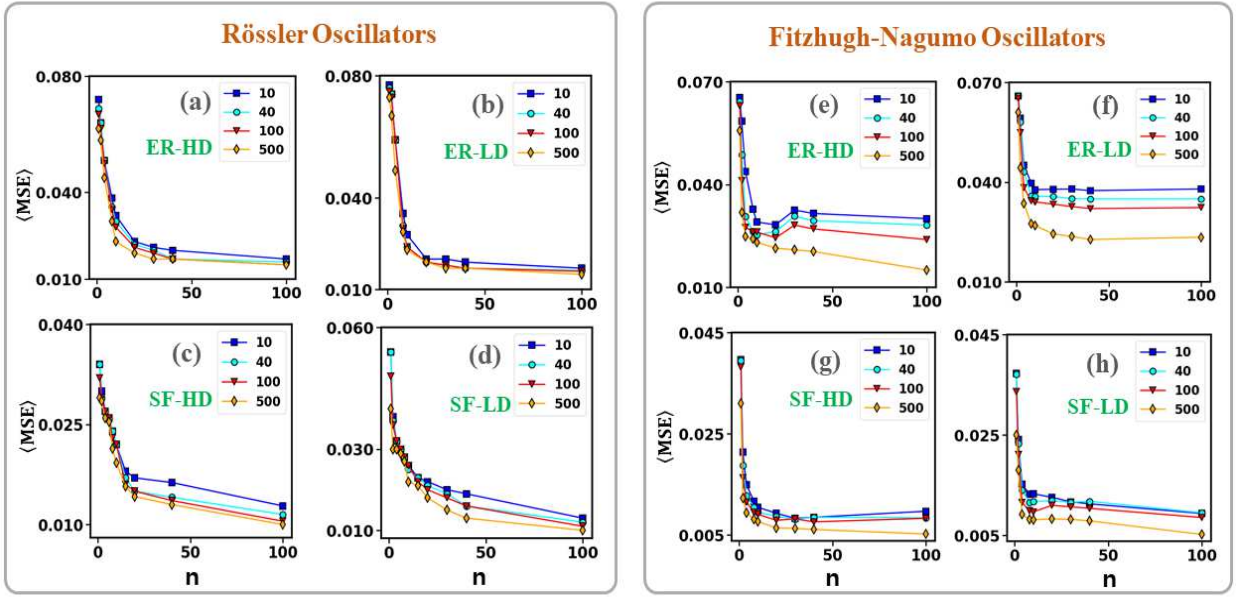}
\caption{Average Mean Square Error ($\langle MSE \rangle$) between true ({\bf R}) and predicted ($\widetilde{\bf R}$) correlation matrices associated with the test data sets by varying $n$ and window size ($w$). Here, $n$ is the number of top degree (HD) or bottom (LD) degree nodes. (a, c) the number of higher degree nodes (ER-HD, SF-HD) vs. $\langle MSE \rangle$ for ER and SF networks on R{\" o}ssler model with varying $w$ and (b, d) number of lower degree nodes (ER-LD, SF-LD) vs. $\langle MSE \rangle$ for ER and SF networks on the R{\" o}ssler model with varying $w$. (e-h) We repeat the same experiment for the FHN model on ER and SF networks. The plots show saturation in $\langle MSE \rangle$ with an increase in $n$, asserting that only limited nodes are required for predicting the entire correlation matrix for both models. For the oscillators, we choose ($\lambda_{R} =0.015$, $\lambda_{F}=0.32$) for ER network realizations, ($\lambda_{R} =0.009$, $\lambda_{F}=0.28$) for the SF network  realizations. We consider $w=\{10,40,100,500\}$ and $skip=\{40,100\}$. }
\label{n_MSE} 
\end{figure}

Since our goal is to predict an entire correlation matrix from the partial information, we roll a window of size $n \times w$ on ${\bf X}^{k} \in \mathbb{R}^{N\times L}$ and creates time series windows ($\Delta{\bf X}_{q}^{k} \in \mathbb{R}^{n \times w}$) such that $n < N$, $w < L$, $1\leq q \leq f$, where $f=\lceil((L-w)/skip)\rceil+1$ is the number of windows, and skip is the gap between two consecutive windows in time series data (Fig. \ref{schematic_train_test} (a) and (c)). Here, $n$ is the number of time series associated with higher or lower-degree nodes we choose to create the windows. Hence, we create $f$ number of windows from each time series data set, ${\bf X}^{k}$. As there are $d$ different time series data sets, we have a total of $fd$ windows for a particular dynamical model. We use these $fd$ windows as the input set to the ML model, which is used as training and testing data sets (Fig. \ref{schematic_train_test} (b) and (d)). The time series window matrix $\Delta{\bf X}_{q}^{k}$ and the predicted correlation matrix ($\widetilde{\bf R}_{q}^{k}$) can be represented as 
\begin{equation}\nonumber
\begin{gathered}
\Delta{\bf X}_{q}^{k}
    =\begin{pmatrix}
\Delta x_{11}  & \Delta x_{12} \dots & \Delta x_{1w}\\
\Delta x_{21}  & \Delta x_{22}  \dots & \Delta x_{2w}\\
\vdots  & \vdots   \ddots & \vdots\\
\Delta x_{n1}  & \Delta x_{n2}  \dots & \Delta x_{nw}
\end{pmatrix},
\Delta \bm{x}_{q}^{k}
    =\begin{pmatrix}
\Delta x_{1} \\
\Delta x_{2} \\
\vdots \\
\Delta x_{nw}
\end{pmatrix},
\widetilde{\bf R}_{q}^{k}
=\begin{pmatrix}
\widetilde{r}_{11}  & \widetilde{r}_{12} \dots & \widetilde{r}_{1N}\\
\widetilde{r}_{21}  & \widetilde{r}_{22} \dots & \widetilde{r}_{2N}\\
\vdots  & \vdots \ddots & \vdots\\
\widetilde{r}_{N1} & \widetilde{r}_{N2} \ddots & \widetilde{r}_{NN}
\end{pmatrix},
\widetilde{\bm{r}}_{q}^{k}
=\begin{pmatrix}
\widetilde{r}_{1}\\
\vdots\\
\widetilde{r}_{i}\\
\vdots\\
\widetilde{r}_{m}
\end{pmatrix}
\end{gathered}
\end{equation}
with elements $\Delta x_{ij}$ where $1 \leq i \leq n$, and $1 \leq j \leq w$.
Further, for the ML model, the input time window matrix ($\Delta {\bf X}_{q}^{k} \in \mathbb{R}^{n \times w}$, $n<N, w<L$) 
are prepared in the shape of a column vector with information of each node, stacked on top of each other and denoted as $\Delta \bm{x}_{q}^{k}\in \mathbb{R}^{nw \times 1}$ (Fig. \ref{schematic_train_test}(b)). Therefore, we assign $nw$ neurons for the input layer of the ML model. Here, $\widetilde{\bm{r}}_{q}^{k}$ represents upper triangular part of $\widetilde{\bf R}_{q}^{k}$.
\begin{table*}[tbh]
\begin{center}
\begin{tabular}{|c|p{5cm}|p{4.5cm}|} 
\hline
{\bf Models}&{\bf R{\"o}ssler} (${\bf \lambda_{R}}$)& {\bf FitzHugh-Nagumo} (${\bf \lambda_{F}}$)\\
\hline &&\\[-1em]
$\mathcal{G}_1^{ER},\mathcal{G}_2^{ER},\ldots,\mathcal{G}_{75}^{ER}$ &$0.012, 0.013, 0.014, 0.015, 0.016$ &$0.28, 0.3, 0.32, 0.34$ \\ 
\hline &&\\[-1em]
$\mathcal{G}_1^{SF},\mathcal{G}_2^{SF},\ldots,\mathcal{G}_{75}^{SF}$&$0.005, 0.006, 0.007, 0.008, 0.009$ & $0.24, 0.26, 0.28, 0.3$ 
\\
\hline
\end{tabular}
\end{center}
\caption{Combination of network models (ER and SF) and chosen coupling strength ($\lambda_{R}$ and $\lambda_{F}$) values for time series data generation. We choose $5$ different coupling strength for the R{\" o}ssler oscillators on $75$ ER (SF) network realizations leading to $375$ time series data sets ($\{{\bf X}^{1},{\bf X}^{2},\ldots,{\bf X}^{375}\}$). Similarly, we choose $4$ different coupling strength for the FitzHugh-Nagumo oscillators on $75$ ER (SF) network realizations leading to another $300$ time series data sets.}
\label{couplings}
\end{table*}
 \begin{figure}[tbh]
\centering
\includegraphics[width=5.5in, height=4.3in]{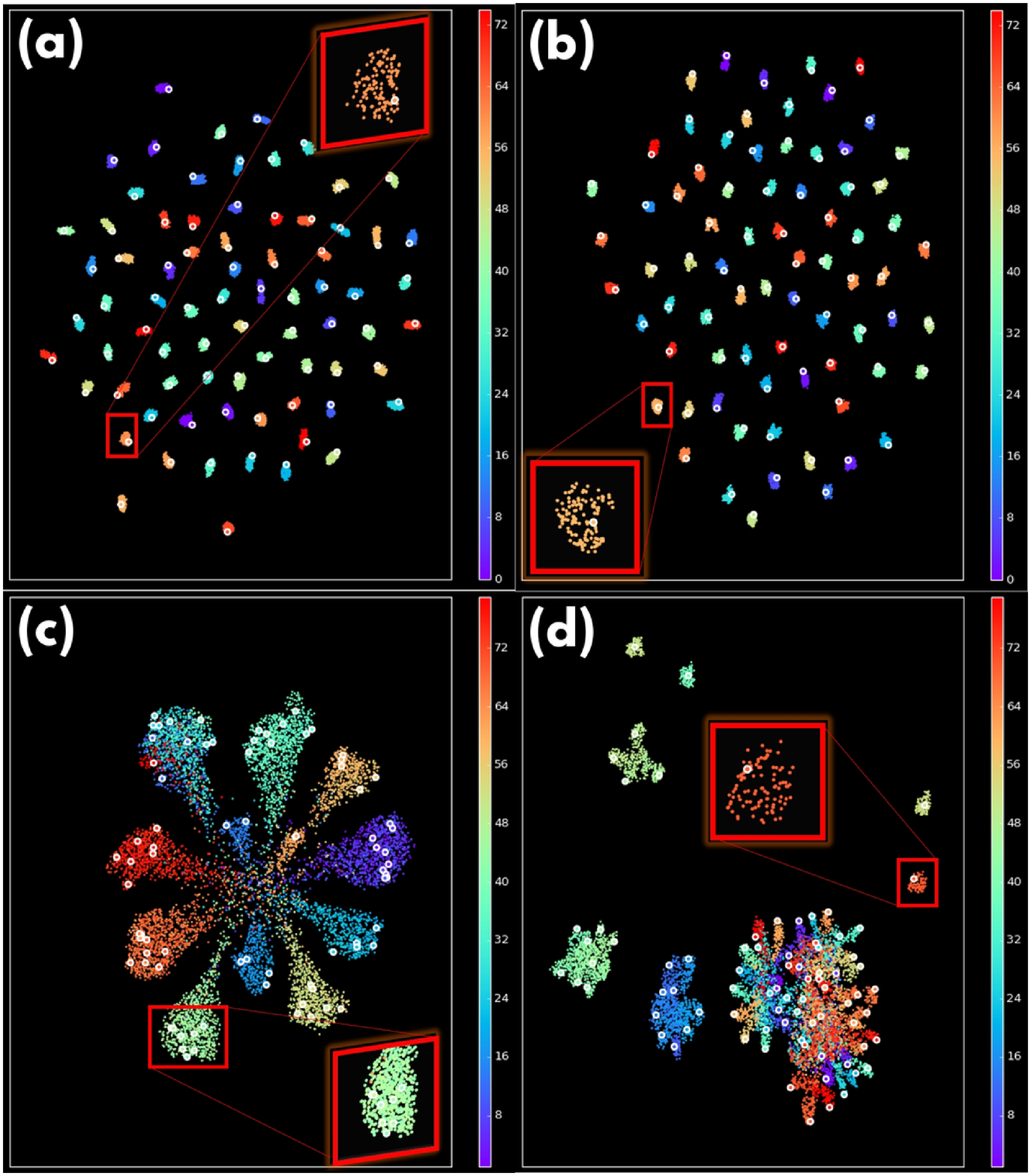}   
\caption{UMAP visualizes the true and predicted correlation matrices as points in 2D space for the test data sets ($\{{\bf X}^{301},{\bf X}^{302},\ldots,{\bf X}^{375}\}$). We consider $N=100$, thus ${\bf R}, \widetilde{\bf R} \in \mathbb{R}^{100 \times 100}$. We take the upper triangular part of a correlation matrix, i.e., $\frac{N^2-N}{2}=4950$ elements, and make a high dimensional vector ($\mathbb{R}^{4950}$) and use UMAP to projects into 2d space ($\mathbb{R}^{2}$). The labeled white color circles represent true correlation matrices ($\{{\bf R}^{301},{\bf R}^{302},\ldots,{\bf R}^{375}\}$) as 2D points. The color cloud dots around the white circle indicate predicted correlation matrices obtained from different time windows associated with a particular time series data set. (a, b) R{\" o}ssler oscillator on ER and SF networks. 
(c, d) FHN oscillator on ER and SF networks. 
For each sub-figure zoomed panel shows the true and predicted correlation matrices as points in 2d space associated with a specific test data set ${\bf X}^{k}$, and Fig. (\ref{stat_distribution}) portrays the corresponding statistical distribution of the correlation matrix elements. }
\label{ER_umap_ross_corr}
\end{figure}
To make the notation simpler, we drop $q$ and $k$ from $\Delta x_{q,ij}^{k}$ and $\widetilde{r}_{q,ij}^{k}$. Note that if we vary $n$, $w$, $skip$, we can create different training and testing data sets.  

\subsection{Prediction of the correlation matrix for varying coupling strengths and network realizations}

For our experiment, we choose $5$ different coupling strengths ($\lambda_{R}$) values for R\"ossler Oscillator on $75$ different ER network realizations ($\mathcal{G}_{1}^{ER}$,$\mathcal{G}_{2}^{ER}$,\ldots,$\mathcal{G}_{75}^{ER}$) to create time series data sets (Table \ref{couplings}).  
Hence, we have $375$  different time series data sets and corresponding true correlation matrices. The true correlation matrices are obtained from $L$ length time evolution of all the nodes. We set the size of the network to $N=100$ and time evolution $L$ set to $5000$, thus ${\bf X}^{k} \in \mathbb{R}^{100 \times 5000}$ and ${\bf R}^{k} \in \mathbb{R}^{100 \times 100}$. We consider training set as ($\{{\bf X}^{1},{\bf X}^{2},\ldots,{\bf X}^{300}\}$, $\{{\bf R}^{1},{\bf R}^{2},\ldots,{\bf R}^{300}\}$) and test set as ($\{{\bf X}^{301},{\bf X}^{302},\ldots,{\bf X}^{375}\}$, $\{{\bf R}^{301},{\bf R}^{302},\ldots,{\bf R}^{375}\}$) for the R{\" o}ssler oscillators on ER network realizations. 

Finally, we create time series windows for each of the data sets (${\bf X}^{k}$). As shown in Fig. \ref{schematic_train_test}(a), we create model inputs by moving a window across the data set in the time-series direction. From $300$  training data sets, we generate $300*124=37200$ windows ($f = \lceil((5000-100)/40)\rceil+1=124$ number of windows, where $L=5000$, $w=100$, $skip=40$) and from $75$ test data sets, we generate $9300$ windows (Fig. \ref{schematic_train_test}(c)). Therefore, training data sets will contain $37200$ windows and $300$ 
true correlation matrices and test data set having $9300$ windows and $75$ true correlation matrices. 
We repeat the time series data sets preparation by varying the oscillator and network models (Table \ref{couplings}).

In the training phase of the supervised ML algorithm, we give the time series windows and true correlation matrices as input to the model. The model adjusts the weights by predicting the correlation matrices from the windows. Using the loss function (Eq. \ref{eq.cost_fnc}), ML model will update the $w_{ij}$ values (Eq. \ref{eq.FCN_layer}) which will minimize the $\mathcal{L}$. The model is trained in the direction of minimizing the difference between the generated correlation matrix and the true correlation matrix (Fig. \ref{schematic_train_test}(a and b)). 

\begin{figure}[tbh]
\centering
\includegraphics[width=4.7in, height=3.6in]{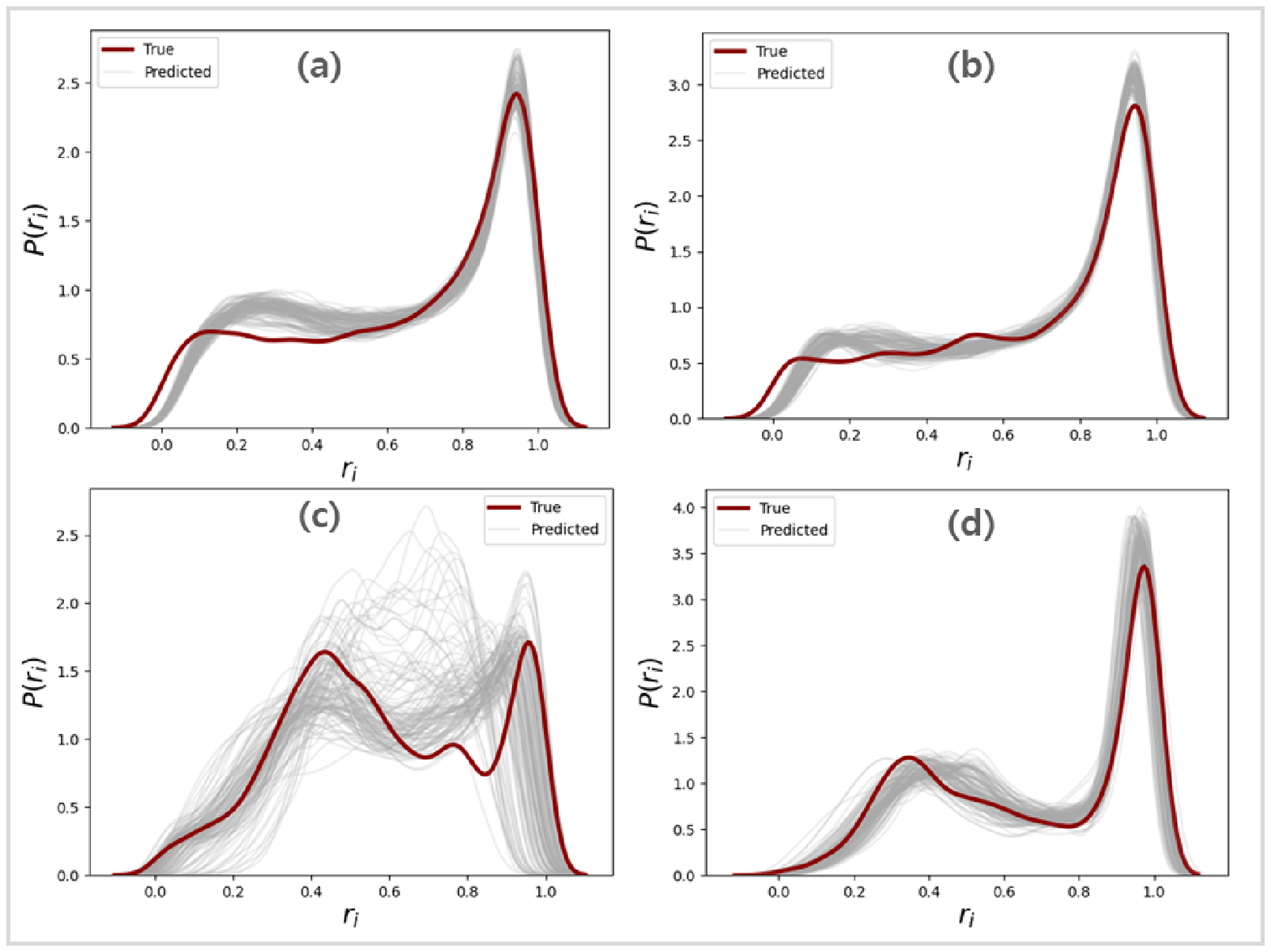}   
\caption{Statistical distribution of the elements of the true and predicted correlation matrices associated with a particular test data set (${\bf X}^{k}$). We take the upper triangular part of a correlation matrix associated with ${\bf X}^{k}$ as $\bm{r}^k=(r_1,r_2,\ldots,r_{m})^{T}$ and prepare a histogram. Finally, we use the Kernal Density Estimation function to make the density plot. We repeat the steps for all the windows associated with ${\bf X}^{k}$. (a, b) R{\" o}ssler oscillators on ER and SF network realizations and (c, d) FHN oscillators on ER and SF network realizations.
We can observe that the distribution of the elements of predicted correlation matrices is very close to the true correlation matrix, which we can also observe from Fig. \ref{ER_umap_ross_corr}. The difference is that UMAP visualizes the correlation matrix as a point and here as a statistical distribution.}
\label{stat_distribution}
\end{figure}
In the testing phase, the model predicts correlation matrices for time series windows not used for training. The supervised learning method takes the inputs of the time series of a few nodes as a window ($\Delta {\bf X}$) and predicts the correlation matrix ($\widetilde{\bf R}$). To evaluate the performance during test time, we compare the predicted correlation matrices ($\widetilde{\bf R}_{q}^{k}$) from the ML algorithm with the true correlation matrix (${\bf R}^{k}$) using Mean Square Error (MSE) measure. Figure \ref{n_MSE} (a) shows the average MSE (Eq. (\ref{avg_mse})) between true (${\bf R}^{k}\rightarrow \bm{r}^{k}$) and predicted ($\widetilde{\bf R}_{q}^{k}\rightarrow \widetilde{\bm{r}}_{q}^{k}$) correlation matrices associated to the test data sets for R{\" o}ssler on ER networks by varying $n$ and $w$. Here, $n$ is the number of top-degree nodes (HD) of the network (Fig. \ref{n_MSE} (a)). For example, if we fix $w=10$ and $n=1$, then time series data associated with the maximum degree node (ER-HD) is considered for window creation. If $n=2$, then time series data associated with two top degree nodes are considered for window creation, and so on. The average MSE can be defined as 
\begin{equation}\label{avg_mse}
\langle MSE \rangle = \frac{1}{f|\mathcal{T}|}\sum_{i=1}^{f|\mathcal{T}|} MSE_{qi}^{k},\text{ where } MSE_{qi}^{k} = \frac{1}{m}\sum_{j=1}^{m} (r^{k}_{j}-\widetilde{r}_{qj}^{k})^2
\end{equation}
where $|\mathcal{T}|$ is the size of test data sets, $ 1\leq q \leq f$, and $ 1 \leq k \leq |\mathcal{T}|$. The observations show that prediction accuracy reaches saturation after increasing the number of nodes beyond a certain point. Importantly, the saturation in accuracy infers that only a limited time series subset is enough to make good correlation matrix predictions. We vary $w$ and repeat the experiment and observe that the results are the same (Fig. \ref{n_MSE}(a)).

Further, we consider $n$ bottom degree nodes and observe that as $n$ and $w$ vary, $\langle MSE \rangle$ decreases and saturates (Fig. \ref{n_MSE}(b)). The prediction accuracy for both cases is observed to be the same (Fig. \ref{n_MSE}(a-b)). We repeat the experiment by varying the SF structure on R{\" o}ssler (Fig. \ref{n_MSE}(c, d)) as well as on the FHN model (Fig \ref{n_MSE}(e-h)) and the pattern remains the same. Thus, the degree of nodes does not impact the prediction of the correlation matrices. It might be a reason that higher and lower degree nodes are similar due to the small-world effect. As the minimum degree of the network is one, and the network is connected, a path exists between a pair of nodes. That is, if we wait for a sufficient time, the information of the entire network can be delivered to the node of the minimum degree. We removed the transient region from the time series; as a result, correlated information of the entire network is accumulated in both high and low-degree nodes. 

Here, we use the MLP, GRU, LSTM, and CNN models to compare the performance of correlation matrix prediction \cite{cho2014learning, gers2000learning, lecun1998gradient}. We use the RNN structure as {\bf GRU + MLP} and {\bf LSTM + MLP}  composed of two recurrent layers (size of $128$ and $256$) and fully connected layers identical to the {\bf MLP}. We use {\bf CNN+MLP} as a model that changes the fully-connected part from the structure of LeNet-5 \cite{lecun1998gradient} to the same structure as {\bf MLP}. Empirically, GRU trains faster and performs better on small-size training data sets than LSTM. 
However, we can observe that MLP, RNN, and CNN all show similar performance. Since MLP is the basic model (Table \ref{benchmarking}), we performed all experiments using the MLP in this study. However, we can also use GRU, LSTM, or CNN for the correlation matrix prediction task.
\begin{table}[tbh] 
\centering
\begin{tabular}[t]{ccccc}
\hline
 &{\bf MLP}& {\bf GRU+MLP}& {\bf LSTM+MLP}& {\bf CNN+MLP}\\
\hline
{\bf MSE}& {\bf 0.0178} $\pm$ & 0.0181 $\pm$& 0.0229 $\pm$& 0.0176$\pm$\\
{\bf R{\"o}ssler on ER network}&($1.07\times10^{-4}$)&($8.59\times10^{-5}$)&($2.53\times10^{-4}$)&($7.39\times10^{-4}$)\\
\hline
{\bf MSE}& {\bf 0.0138} $\pm$& 0.0138$\pm$ & 0.0188$\pm$& 0.0140$\pm$\\
{\bf R{\"o}ssler on SF network}&($1.54\times10^{-4}$)&($1.71\times10^{-4}$)&($1.87\times10^{-4}$)&($3.26\times10^{-4}$)\\
\hline
\end{tabular}
\caption{Experimental results of different ML models. MSE of the models for the time window with size $n = 20$ and $w = 100$. The standard deviation is indicated in brackets.}
\label{benchmarking}
\end{table}

\subsection{Unsupervised learning method to understand correlation matrix prediction}

We use a dimensionality reduction tool (UMAP) to visualize and understand how close the true and predicted correlation matrices are obtained from the testing phase of the supervised ML model. The UMAP helps to understand the similarities between predicted ($\widetilde{\bf R}$) and true correlation ({\bf R}) matrices by considering whole matrices as points in high dimensional space and embedding them in 2D space. 

For instance, we consider $9300$ windows of the $75$ test data sets associated with  R{\"o}ssler oscillators on different ER network realizations. The UMAP algorithm takes the upper triangular part of all flattened correlation matrices (predicted and true) corresponding to the test data sets as a high dimensional input vector and projects them as points in the lower dimensional space. 

From Fig. \ref{ER_umap_ross_corr}(a), one can observe that $75$ different clusters correspond to true and predicted correlation matrices as points in 2D space for $75$ different test time series data sets. Further, all true correlation matrices are marked with white-colored circles, and the predicted correlation matrices form a cloud around the true correlation matrix. Importantly, the predicted and true correlation matrices for a specific ${\bf X}^{k}$ are close in the 2D space (Fig. \ref{ER_umap_ross_corr}(a)) and
are distributed only near the corresponding true correlation matrix (Fig. \ref{stat_distribution}(a)), inferring that the predictions made are meaningful. The SF network realizations on R{\"o}ssler oscillator also show similar behavior (Figs. \ref{ER_umap_ross_corr}(b) and \ref{stat_distribution}(b)).

Further, one can notice that for the FHN oscillators on the ER and the SF network realizations, unlike the case of the R{\"o}ssler oscillator, predictions on them do not constitute isolated clusters with true correlation matrices (Fig. \ref{ER_umap_ross_corr}(c) and (d)). However, we can still observe that the predicted correlation matrices are located near the true correlation matrices in the UMAP and distribution plots (Fig. \ref{stat_distribution}(c) and (d)). The fact that true correlation matrices constitute a cluster means they share similar characteristics.
Therefore, we can assert that the model makes meaningful predictions (Figs. \ref{ER_umap_ross_corr} and \ref{stat_distribution}).

\begin{figure}[tbh]
\centering
\includegraphics[width=6.5in, height=4.1in]{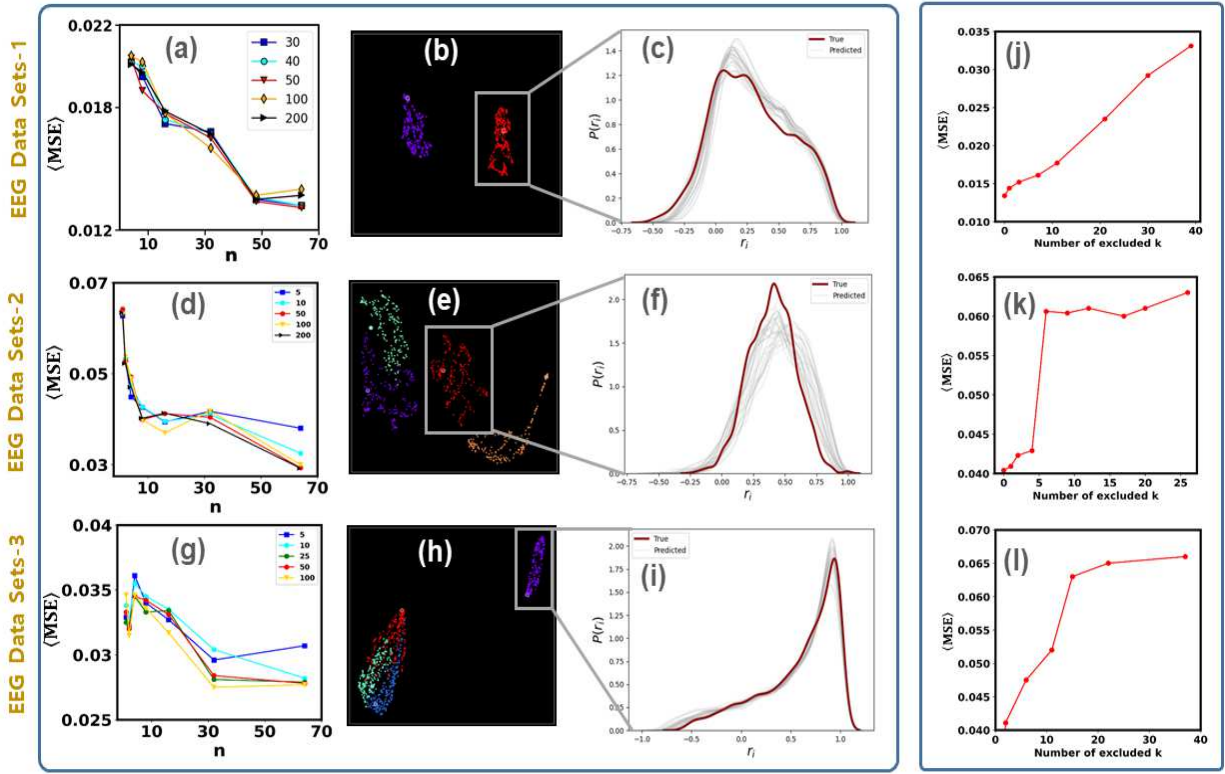}
\caption{(a) Average Mean Square Error ($\langle MSE \rangle$) between true ({\bf R}) and predicted ($\widetilde{\bf R}$) correlation matrices for EEG test data sets \cite{liu2020beta} as number of channels ($n$) and window size vary. We observe decreasing $\langle MSE \rangle$ with an increase in $n$. Asserting that only a limited number of channel's EEG data are required to predict the entire correlation matrix. (b) UMAP is generated from the {\bf R} and $\widetilde{\bf R}$ of the EEG test data sets. Among the $65$ subjects ($\{{\bf X}^{k}\}_{k=1}^{65}$) in EEG data, two subjects use as the test data set. The labeled color indicates two different subjects, while white circles represent the true correlation matrix. The color dots embedded points from $\widetilde{\bf R}$s corresponding to different windows. The $\widetilde{\bf R}$s created from $\Delta {\bf X} \in \mathbb{R}^{n \times w} (n=48, w=50)$. (c) shows the distribution plot of a cluster. (d-i) We repeat the same experiment for another two EEG data sets with $64$-channel and $109$ subjects \cite{goldberger2000physiobank, schalk2004bci2000}. Here, $4$ data sets are used for testing and observing that model can predict the correlation matrices. (j-l) Finally, we performed the ablation study by reducing the training data set size and training the model. For all the data sets, one can observe that reducing train data sets increases the $\langle MSE \rangle$ in the test phase of the model.}
\label{EEG} 
\end{figure}

\subsection{Experiment on EEG Data}
To validate our model, we use the brain-computer interface data sets in our study \cite{liu2020beta, goldberger2000physiobank, schalk2004bci2000}. The first database comprises $64$-channel Electroencephalogram (EEG) data of $70$ subjects performing a 40-target cued-spelling task. The EEG data are stored as a 4-way tensor, with a dimension of channel $\times$ time point $\times$ block $\times$ condition. Our experiment considers time point vs. channel data for $65$ subjects of the first block and condition one. Hence, we have $65$ different time series data sets ($\{{\bf X}^{1},{\bf X}^{2},\ldots,{\bf X}^{65}\}$) each of having $N=64$ time series and length, $L=750$. From the time series data sets, we create the corresponding true correlation matrices and denoted as ($\{{\bf R}^{1},{\bf R}^{2},\ldots,{\bf R}^{65}\}$).
Among the $65$ EEG data sets, we use $63$ as training data and $2$ data sets used as test data. The training and test sets division are the same as predicting unknown realization in the  R{\"o}ssler and FHN experiments. Further, we create $n \times w$ size windows for the training and test data sets. For instance, for any ${\bf X}^{k} \in \mathbb{R}^{N \times L}$ we have $f=\lceil (L-w)/skip\rceil+1=351$ windows ($\Delta {\bf X}^{k}_{q} \in \mathbb{R}^{n \times w}$, $1 \leq k \leq 65$, $1 \leq q \leq 351$) where $n=48$, $w=50$ and $skip=2$. Hence, the number of windows for training data sets is $351*63=22113$ and $351* 2=702$ for the test data sets. We train the model using $22113$ windows and $63$ true correlation matrices. During the test phase, we use a window in test data sets to predict the correlation matrix ($\widetilde{\bf R}$). We vary $n$ and $w$ to create other training and test data sets and repeat the experiment. As shown in Fig. \ref{EEG} (a), the average MSE decreases as $n$ increases. But for $w$, it does not affect the performance much until $w=200$. The $\langle MSE \rangle$ converges around $n=48$. If we look at the UMAP and the distribution for $\widetilde{\bf R}$ and {\bf R}, we can see that $\widetilde{\bf R}$ is distributed near the true {\bf R} (Fig. \ref{EEG}(b, c)). 

We use another two data sets consisting of $64$ channels of EEG recordings, obtained from $109$ subjects performing different motor/imagery tasks (Baseline, eyes open) and  (open and close left or right first), respectively \cite{goldberger2000physiobank, schalk2004bci2000}. Hence, we have $109$ time series data sets, each having length $L=2000$. Among the $109$ EEG data sets, we use $105$ as training data sets and $4$ as test data sets. One can observe that the average MSE decreases as $n$ increases and converges around $n=20$ (Fig. \ref{EEG}(d)). Again, if we look at the UMAP and distribution for $\widetilde{\bf R}$ and {\bf R}, we can see that $\widetilde{\bf R}$ is distributed near the true {\bf R} (Fig. \ref{EEG}(e, f)). Similarly, we can observe the same predictive nature for other data sets (Figs. \ref{EEG}(g-i)). In other words, it can be seen that the ML model predicts $\widetilde{\bf R}$, reflecting each subject's unique characteristics. 
It suggests that the model predicts the correlation matrix in the EEG data set. With the experimental results obtained from the model and EEG data sets, the framework can be applied to other real-world data sets. 

Finally, we perform the ablation study by successively reducing the training data set size and training the ML model. It helps us to understand the impact of training data sets size on the error during the test phase. We can observe that during the test time, $\langle MSE \rangle$ errors increase as we remove more training data sets (Fig. \ref{EEG}(j-l)). For certain data sets, reducing makes increasing the $\langle MSE \rangle$ and then saturates. Our UMAP analysis reveals that the error increases significantly when a train data set similar to the test set is excluded. But when train data sets relatively less similar to the test set are excluded, the test error is not significantly affected. We refer to the UMAP figures on EEG data sets in SI for more details. Overall, we can say that increasing training data sets up to some limit will improve the model performance. The model may sometimes fail when the true correlation matrix of the test data sets is far away from all the true correlation matrix of the training data sets.

\section{Conclusion}
We present a framework that combines supervised and unsupervised learning to predict the correlation matrix of the entire system from limited time series data available for a subset of nodes. We use two well-known chaotic oscillator systems for time series data generation by choosing the appropriate coupling strengths. In addition, we use both linear (Pearson) and nonlinear (Spearman) correlation functions to measure the correlation between a pair of time series data of nodes. We observe that for both cases, the results are the same.

Supervised learning has been applied to make predictions from limited time series windows. Its prediction quality has been measured using mean square error, the difference between the true and the predicted correlation matrix. The threshold of the number of nodes required and the length of the limited time series to accomplish good predictions have also been discussed. The correlation matrices are predicted using time series data associated with higher-degree or lower-degree nodes. The prediction accuracy indicates that the degree of the available nodes associated with a time series data does not impact the correlation matrix prediction. After that, unsupervised learning (UMAP) brought more insights into the prediction results by visualizing the results. Finally, we examine real-world EEG data sets to validate our model. We also incorporate an ablation study for the error analysis to understand the model's usefulness in real-world data sets.

We currently focused on the EEG data sets since they have a similar structure as our modeled data sets. Our model provides good correlation matrix prediction for all the used real-world data sets. However, we can use our model on other data sets in the future. Here, most results are generated using the MLP model, and we use other ML models for comparative study. However, we can use more advanced ML architectures such as GRU, LSTM, or CNN for the correlation matrix prediction task for larger network structures and which require further investigation.

There is evidence of oscillations and chaos in neural networks \cite{wang1990oscillations}, and recent studies by neurologists found chaos in the human brain \cite{Hamzelou2023}. However, using chaotic systems to generate time series data and using the NN model to predict the correlation matrix from the partial time series data is new. Our model provides good prediction results for both modeled and real-world data sets. Our work may open up a window into an area of ML in complex networks where predictions can be made possible by using only a finite length of time series data over a finite number of network nodes.


\section*{Acknowledgment}
NE and PL are thankful to Chittaranjan Hens (IIIT Hyderabad) for the useful discussion on the FHN model. PP is indebted to Kritiprassna Das (IIT Indore) for a detailed discussion on the EEG datasets. SJ acknowledges DST grant SPF/2021/000136.





\end{document}